\documentclass[letterpaper, 10 pt, conference]{ieeeconf}
\usepackage{graphicx}
\usepackage{algorithm}
\usepackage{algpseudocode}
\usepackage{multirow}
\usepackage{hyperref}
\usepackage[inkscapelatex=false]{svg}
\usepackage[utf8]{inputenc}
\IEEEoverridecommandlockouts
\title{\LARGE \bf
Inferring World Belief States in Dynamic Real-World Environments
}

\author{Jack Kolb, Aditya Garg, Nikolai Warner, and Karen M. Feigh
\thanks{We thank Danielle Generous for her artistic contributions.}
\thanks{All authors are with the College of Engineering at the Georgia Institute of Technology, North Avenue, Atlanta, GA 30332, USA {\tt\small \{kolb, agarg354, nwarner30, karen.feigh\}@gatech.edu}.}
}

\begin{document}
\maketitle
\thispagestyle{empty}
\pagestyle{empty}

\begin{abstract}
We investigate estimating a human's world belief state using a robot's observations in a dynamic, 3D, and partially observable environment. The methods are grounded in \textit{mental model} theory, which posits that human decision making, contextual reasoning, situation awareness, and behavior planning draw from an internal simulation or world belief state. When in teams, the mental model also includes a \textit{team model} of each teammate's beliefs and capabilities, enabling fluent teamwork without the need for constant and explicit communication. In this work we replicate a core component of the team model by inferring a teammate's belief state, or \textit{level one situation awareness}, as a human-robot team navigates a household environment. We evaluate our methods in a realistic simulation, extend to a real-world robot platform, and demonstrate a downstream application of the belief state through an \textit{active assistance} semantic reasoning task.
\end{abstract}

\section{INTRODUCTION}

The cognitive engineering community holds that people represent their surroundings via an internal simulation termed a \textit{mental model} \cite{endsley1988situation}. The mental model contains all task-relevant information the person is aware of, such as the locations of environment objects, higher-level semantic relationships (i.e. context), and the current task goals \cite{cannon1993shared}. Downstream human factors such as situation awareness and behavior planning are directly informed by the mental model.

In human-human teams, mental models include a \textit{team model} that includes team goals, role assignments, and predictions of the mental models of teammates \cite{tabrez2020survey}. The team model is used to inform a person's actions, planning, and communication contextualized by their understanding of the teammates and team objectives. Practically, the team model facilitates important aspects of team dynamics including theory of mind, implied objectives, non-verbal coordination, and selective communication.

\begin{figure}[t!]
    \centering
    \includegraphics[width=0.996\linewidth]{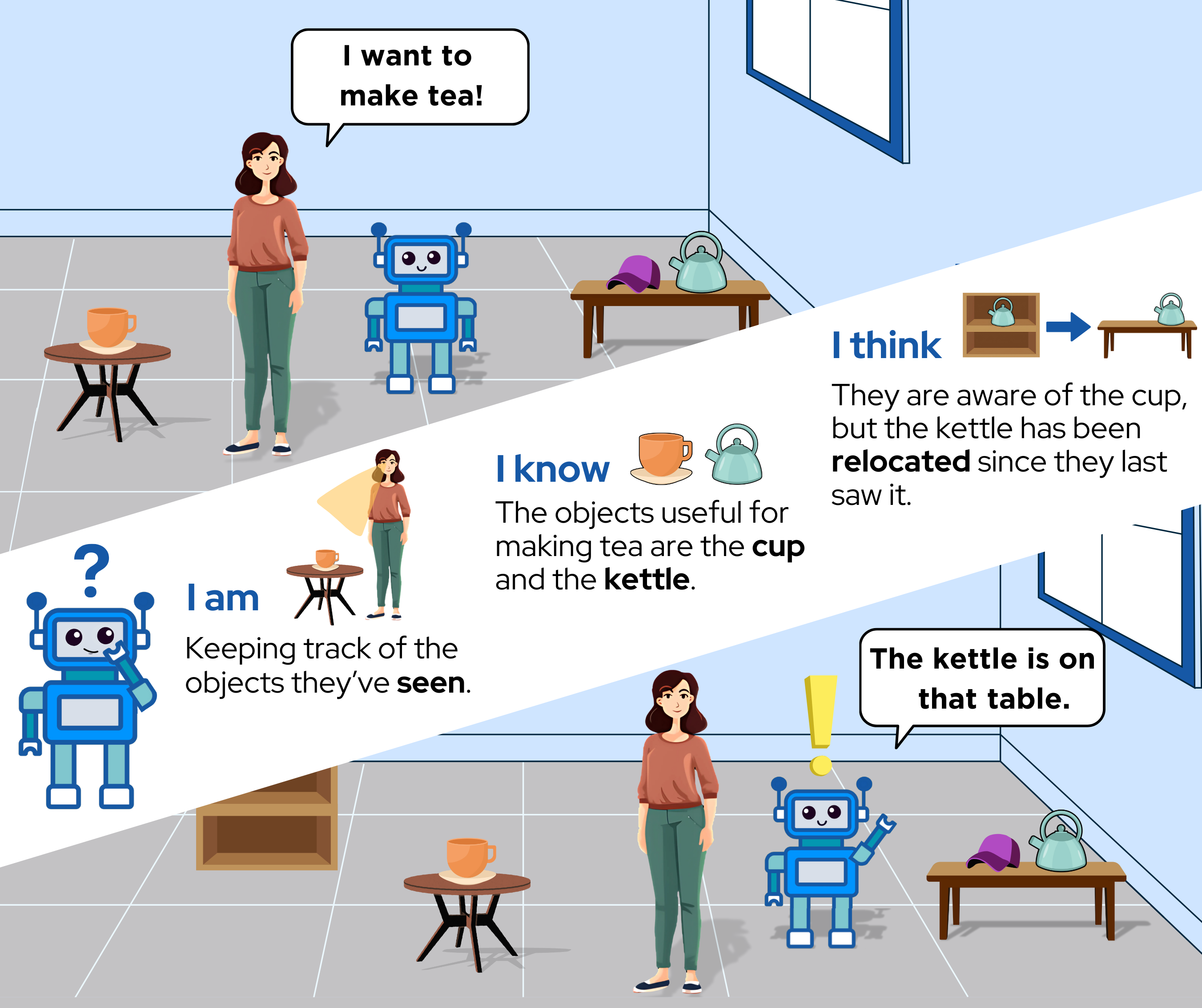}
    \caption{Overview of the predicted user belief state system. The robot observes a person in their environment and estimates their current belief state. The belief state can be applied to various downstream reasoning tasks.}
    \label{fig:overview}
\end{figure}  

This work answers the research question: \textit{Can a robot accurately infer the belief state of a person in the environment?} We define a belief state as \textit{level one situation awareness} \cite{endsley1988situation}, or the person's understanding of environment objects and their locations. Specifically, we are interested in human-centered spaces with large floor plans such as households, storefronts, or industrial environments.

Building upon prior work which explored a simplified variant of this problem \cite{kolb2024inferring}, we present a first approach and evaluate our methods in a simulated household domain. A robot and a person navigate the house, and the robot continuously updates its model of the person's belief state from sparsely observing the person's location and pose.

The domain introduces four key challenges:

\begin{enumerate}
    \item \textbf{Partial-Observability}: The robot's field of view is constrained to a mounted camera, instead of full observability over the household (e.g., a smart home).  
    \item \textbf{Dynamic Environments}: Objects can be relocated, and people can move and occlude objects. 
    \item \textbf{Object Permanence}: Objects must be remembered and resolved, even when relocated, to closely represent the current world state.
    \item \textbf{Portability}: The explicit belief state can be used in downstream tasks (e.g., theory of mind reasoning) or as a component in larger systems.
\end{enumerate}

This work is the first to address all four challenges \textit{without} relying on privileged information. Using an RGB/D camera stream and robot localization, the methods maintain the robot's own belief state and are used to infer the person's belief state using an explainable and portable data structure. Moreover, the methods are transferrable to other domains.

We demonstrate a downstream use case of our methods by applying the inferred belief state to \textit{active assistance} (see Fig. \ref{fig:overview}). We construct a scenario where a person enters their household to discover that many objects were moved around. The person walks through the household to assess the current state, and then desires to start an activity (e.g., cooking or cleaning). A robot following the person -- and inferring their belief state -- identifies and communicates the set of objects that are unknown to the person and relevant to the task.
\begin{figure*}
    \centering
    \includegraphics[width=1.0\linewidth]{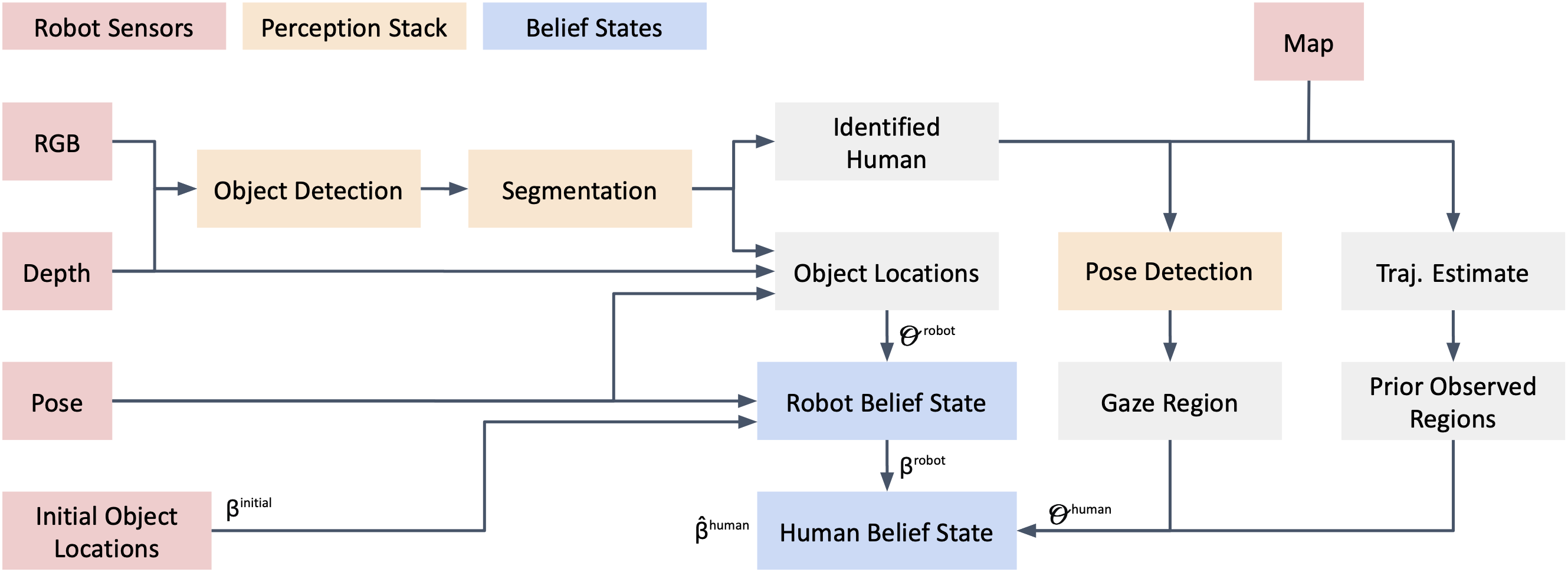}
    \caption{Overview of the predicted user belief state system. The ground truth world state information is filtered by the robot's visibility ($R^{robot})$ and used to update the robot's belief state, $\beta^{robot}$. The robot's belief state is then filtered by the user's visibility to update the predicted user belief state, $\beta^{pred}$. The resulting belief state is what the robot thinks the user is aware of, i.e., a \textit{theory of mind} that can inform downstream reasoning tasks.}
    \label{fig:system}
\end{figure*}
\section{BACKGROUND}

Many open research questions exist for monitoring a person's situation awareness to the detail required for real-time human-AI teaming. The human factors community primarily uses two tools to measure user situation awareness: the SAGAT technique \cite{endsley1988situation} where the task is periodically paused and the user is queried about environment elements and their contextual meaning, and SART-like after-action surveys to gauge situation awareness during the preceding task \cite{selcon1990evaluation}.

For situation awareness monitoring to be deployed to real-world domains, robot systems would require \textit{passive} assessment with minimal intervention. Additionally, inferring situation awareness from non-intrusive third party observations is preferred over relying on wearable devices such eye-level cameras, gaze tracking glasses, or EEG headsets. 

Related works have inferred situation awareness using gaze detection, by conducting post-hoc classification or training a saliency map madel. Such works target specific task domains, such as driving or air traffic control, where users focus on a defined task in a small workspace \cite{bhavsar2017quantifying, biswas2024modeling, li2023recognising}. We are instead interested in domains where people are ambulatory and not necessarily fixated on a small region.

Other work has proposed explicit representations that can be applied to many human-AI teaming domains. We observe two paradigms in this space: \textit{latent representations} that embed the belief state within an end-to-end architecture, and \textit{explicit representations} that model the user's awareness of environment objects in a defined and portable format.

The representations present a tradeoff. Latent representations are tailored to specific inference goals and can model abstract user preferences. For example, recent research has sought to estimate a person's belief state while driving, which has primarily relied on training a model to construct a saliency map from a camera located near the driver's perspective \cite{biswas2024modeling}. However, these end-to-end approaches have limited use outside of the model's specific domain objectives.

Alternatively, explicit representations can inform a variety of downstream inference and reasoning tasks. However, they are challenging to define, model, and reliably construct. Several recent works have proposed frameworks for constructing team models with explicit representations \cite{bolton2022fuzzy, edgar2023improving, scheutz2017framework, kolb2024inferring}. We are specifically interested in explicit representations as their data structures are portable to downstream use cases.

All prior works constructing explicit representations have been relegated to abstracted simulated environments. Scheutz et al. proposed a framework built on rules-based \textit{logical predicates} \cite{scheutz2017framework, gervits2020toward}. Applied to a simulated space station repair scenario, two robots constructed team mental models of each other and a human worker to coordinate actions.

Bolton et al. proposed an uncertainty framework built upon three functions -- the user's perception (\textit{what did the user perceive?}), comprehension (\textit{how did the user interpret the event?}), and belief state update (\textit{how did the event change the user's belief?}) \cite{bolton2022fuzzy}. While interesting, the functions are challenging to define and implement in real-world domains.

Our own prior work used a logical predicates framework to infer a person's belief state in a 2D partially-observable cooking domain, and evaluated several approaches to predicting user responses to situation awareness questions \cite{kolb2024inferring}.

No works have approached realistic 3D domains, which introduce technical challenges and are essential for deployment to the real world. Moreover, for a robot or embodied AI system to infer the belief state of a person, the robot first requires its own accurate belief state (i.e., a semantic map), for which virtually no software packages exist. One candidate is the SLAM package Khronos \cite{schmid2024khronos}, however its SLAM focus prioritizes real-time mapping over object permanence.

Notably, the community lacks a software package for constructing a semantic map that supports partial observability, dynamic environments, \textit{and} object permanence. In this work we present a baseline algorithm that supports all four challenges, and future work can improve upon its robustness.

\section{PROBLEM STATEMENT}

We define a belief state $\mathcal{\beta}$ as a set of environment objects $\{o\}$ and their properties (e.g., location, attributes). The belief state is analogous to \textit{level one situation awareness}, or an agent's understanding of objects in the workspace and their semantic properties. While contextual information between objects can be inferred, it is outside our problem scope.

Our objective is for a robot or other embodied AI system to infer the belief state $\mathcal{\beta}^{human}$ of a person in the environment. The robot only has access to traditional sensor information including an initial map of the environment $\mathcal{\beta}^{initial}$, an RGB/D camera, and the robot's current location.

The environment is dynamic (objects can be relocated) and partially-observable (the robot cannot see the entire workspace at once), representative of a household, warehouse, office, or other human-centered domain. Therefore, the robot must be capable of \textit{object permanence} by resolving \textit{which} objects it observes in the environment.

To evaluate the accuracy of the inferred belief state $\mathcal{\widehat\beta}^{human}$ to the person's belief state $\mathcal{\beta}^{human}$ we propose a set distance metric termed SMCC (Summed Minimum Cost by Class). SMCC considers objects within a class as interchangeable, e.g., if a person thinks a spoon is at a location it does not matter \textit{which} spoon is there. Consequently, an object that is important enough to matter at an instance-level should have its own class.

SMCC is defined by Eq. \ref{SMCC} below:

\begin{equation}
SMCC(A, B) = \sum_{c \in C}{minCost(A_c, B_c)}   
\label{SMCC}
\end{equation}

where $minCost(A_c, B_c)$ is the shortest distance to map the locations of objects of class $c$ from set $A$ to set $B$.

SMCC is a commutative \textit{``lower is better''} metric with units of the unit length. In our reporting we divide SMCC by the quantity of objects in the scene, such that a mean SMCC of $1$ represents an average belief state disparity of $1$ meter per environment object.

The problem formulation presents three figures of merit:

\begin{enumerate}
    \item $SMCC(\widehat\beta^{human}, \beta^{human})$, the error of the inferred belief state with respect to the person's actual belief state, i.e., the \textbf{inference performance}.
    \item $SMCC(\beta^{human}, \beta^{true})$, the error of the person's belief state with respect to the \textit{true} world state; a high SMCC indicates the environment is sufficiently complex for the person to have false beliefs.
    \item $SMCC(\beta^{robot}, \beta^{true})$, the accuracy of the robot's own belief state; a high SMCC indicates the robot struggles to maintain an accurate semantic map and may misinform people in downstream tasks.
\end{enumerate}

We present an architecture and implementation for constructing $\widehat\beta^{human}$ given the problem formulation. While our evaluations use one person in the environment, and we use SMCC only in a simulation environment, the problem statement and methods are scalable. In the real-world a person's belief state can be approximated from a wearable camera or obtained from asking people to map known objects.
\section{METHODS}

We approach the problem using recursive theory of mind. The robot initializes its own belief state $\beta^{robot}$ and the inferred human belief state $\widehat\beta^{human}$ with an initial belief state $\beta^{initial}$ mapped \textit{a priori}. The robot updates $\beta^{robot}$ as it observes objects. When the person is visible, the robot estimates the person's trajectory since their last observation, and then estimates which objects from $\beta^{robot}$ the person would have seen along the route to update $\widehat\beta^{human}$. The algorithm maintains the two belief states indefinitely.

Our implementation has two core subsystems -- a belief state algorithm and a perception stack. Fig. \ref{fig:system} shows a system diagram of the proposed architecture and its required inputs.

\subsection{Belief State Algorithm}

We represent a belief state $\beta$ as a semantic map consisting of a set of objects $\{o\}$, their mean point cloud locations $(o_x, o_y, o_z)$, and their classes $o_c$. Our belief state algorithm takes as input a list of observed objects $\mathcal{O}$ from the perception stack and resolves which object instances are seen.

To prevent the quantity of objects from exploding, we assume an initial semantic map $\beta^{initial}$ is provided \textit{a priori} and that the robot is notified when objects are added to or removed from the environment. This is necessary to enforce a 1:1 mapping. The assumption is rationalized as a robot (re)mapping the environment when it is in a static state (e.g., all people are away or asleep), although future works can explore methods for judging whether an observed object has been newly introduced to the environment.

Algorithm \ref{alg:belief state} shows how a belief state $\beta$ updates given each observed object $o\in\mathcal{O}$ of class $o_c$ at location $(o_x, o_y, o_z)$. The complete implementation of the belief state algorithm is available as a Python library from the project webpage.

\begin{algorithm}
\caption{Updating a belief state from observed objects.}
\label{alg:belief state}
\begin{algorithmic}
\Require Sets $\mathcal{O}_c$, $\beta_c$ for an object class $c$
\Ensure Mapping \( f: \mathcal{O}_c \to \beta_c \) that minimizes total distance
\State Find \( f: \mathcal{O}_c \to \beta_c \) such that
\[
\sum_{o \in \mathcal{O}_c} d(o, f(o))
\]
is minimized, where \( d(a, b) \) is the $L^2$-norm distance.
\State \(\beta_c \gets f(\mathcal{O}_c)\)
\end{algorithmic}
\end{algorithm}

\begin{figure*}
    \centering
    \includegraphics[width=1.0\linewidth]{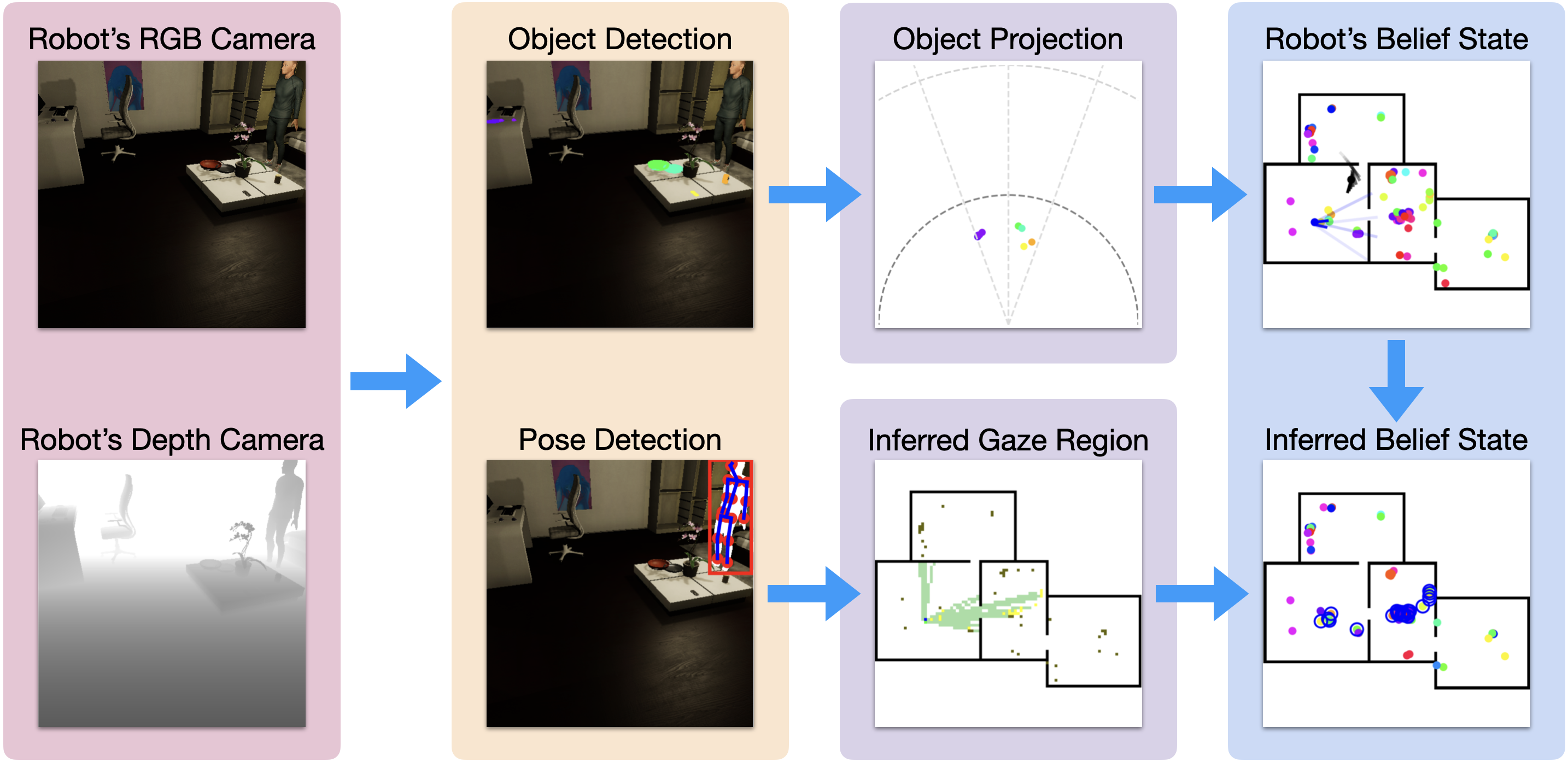}
    \caption{Outline of the perception stack. The robot's RGB camera and depth sensor are used to detect objects in 3D space and resolve the robot's own world belief state $\beta^{robot}$. When a person is observed, their current belief state $\widehat\beta^{human}$ is inferred using their gaze region and estimated prior trajectory.}
    \label{fig:simulation}
\end{figure*}

\subsection{Perception Stack}

The belief state algorithm takes as input a set of observed objects in the environment, each represented by a location and class label. We use OWLv2 \cite{minderer2024scaling} for open-vocabulary object detection, and SAM2 \cite{ravi2024sam} for image segmentation of detected objects. Open-vocabulary detection is chosen as it enables detailed custom classes (e.g., treating the ``blue cup'' as unique from the ``red cup''), automatic generation of relevant objects (e.g., using a vision-language model to identify which objects are useful for certain tasks), and a domain-agnostic implementation. We use a detection confidence threshold of $0.4$ for objects and $0.1$ for people.

A detected object's relative location is computed by masking the depth point cloud with the object's image segmentation to obtain the mean depth of the visible surface and the object's angle relative to the robot's forward vector. The relative location is then projected into global coordinates using the robot's pose and a pinhole camera projection.

Our architecture considers people as distinct from regular objects. As the objective is to infer a person's belief state, false positive and false negative person detections can be consequential. We reduce false negatives by selecting a low detection confidence threshold, inducing the object detector to return more detection candidates. False positives are removed by comparing detections from the RGB image with detections from the depth image, and identifying matches using their bounding box overlap. We use a minimum overlap threshold of $30\%$, and the combined approach virtually eliminated erroneous human detections in our evaluations.

The gaze direction of a detected person is approximated by the forward vector from their 3D pose. RTMPose is used to obtain 2D skeleton keypoints (COCO format) \cite{jiang2023rtmpose}, which are lifted to 3D using MMPose \cite{mmpose}. As accurate head orientation can be unreliable at room-length distances, we estimate the person's gaze direction as the normal vector from the person's hip and shoulder joints. Similar to objects, the pose is projected into global coordinates. 

\subsection{Recursive Theory of Mind}

The inferred belief state \(\widehat\beta^{human}\) uses the same belief state algorithm deployed by the robot. Instead of using the perception stack to identify the observed objects \(\mathcal{O}\), objects are filtered in from the robot's belief state \(\beta^{robot}\) by inferring the person's observable gaze region and trajectory.

Observed objects are identified in two steps. First, candidate objects are selected using the person's location, gaze direction, and approximate field of view. Candidate objects are then raycasted to from the person's location, checking for map collisions using an 8-directional occupancy grid. The resulting objects are passed into Algorithm \ref{alg:belief state} to update \(\widehat\beta^{human}\). Our implementation only considers walls in the occupancy grid, as object occlusions were uncommon from the vantage point of the agents.

To support sparsely observing people, the system tracks the last observed location of the person and models the person's trajectory between observations. We use A* to determine a possible path between observations and update \(\widehat\beta^{human}\) at each occupancy grid cell along the path. 

Future work can improve the trajectory estimate module by using path prediction that models human movement, by calculating object occlusions, and by considering historical object locations when computing the set of observed objects.

\subsection{Simulation Environment}

The presented system is designed to be cross-compatible across real-world sensor inputs \textit{and} simulation inputs. For the quantitative evaluation we use a realistic household simulation environment, VirtualHome \cite{puig2018virtualhome}. The simulator enables reproducible results, ground truth world state information, access to the human agent's perception, and scalable data collection. VirtualHome also natively supports semantic rearrangement tasks and contains a built-in navigation stack. However, the simulator is prone to execution errors, limiting scenarios to approximately five agent actions, or around $90$ seconds of time. This narrows the scope of our evaluations.

VirtualHome is also limited by not providing access to the first-person camera's 6-DOF pose. Instead, the pose must be approximated from the character's skeleton and does not follow character animations. To determine the robot's camera pose we follow the same approach used to approximate the location and gaze direction of an observed person.

We evaluate our system on a scenario termed \textit{``Parents are Out!''} In the scenario, a person leaves the household in a tidy state. While the person is gone, most objects in the house are shuffled around (i.e, the kids throw a party). The person returns to find the house a disaster, assesses the household via a quick walk through, and then desires to do an activity (e.g., cooking, cleaning, or mixing a strong drink). A robot trailing the person aims to infer the person's belief state \(\widehat\beta^{human}\) throughout the scenario's \(\approx 60s\) duration. The robot uses the perception stack (no privileged information) and the human uses ground truth perception from the simulator.

\subsection{Demonstration: Active Assistance}

The robot can reason over \(\beta^{robot}\) and \(\widehat\beta^{human}\) for downstream tasks. We apply this capability to an \textit{active assistance} task, where the robot notifies a person of objects that are relevant to a given task and of which the person has an incorrect belief. Fig. \ref{fig:overview} visually describes the task.

The robot first infers the set of objects that the person has incorrect beliefs of. The resulting set \(\mathcal{O}^{unaware}\) includes all objects in \(\beta^{robot}\) where the object exceeds an \(L^2\)-norm distance threshold from any object of its class in \(\widehat\beta^{human}\).

The robot then prompts an LLM with the activity and \(\mathcal{O}^{unaware}\) to select the subset of objects relevant to the activity. We sought a small, locally-hosted LLM for privacy and deployability on mobile robots, and chose qwen2.5-32B \cite{yang2024qwen2} for its open availability and benchmark performance.

The LLM outputs a set of objects \(\mathcal{O}^{notify}\) to notify the person about. We evaluate three prompt methods: 
\textit{list chain-of-thought} (CoT) queries with the full \(\mathcal{O}^{unaware}\), \textit{single CoT} queries each object individually, and \textit{single binary} queries each object with no chain-of-thought prompting. We also compare against a zero-shot deBERTa-v3 classification model \cite{laurer2024less}. As ground truth, \(\mathcal{O}^{notify}\) is compared against a hand-crafted map from activities to relevant object classes.

The evaluation includes nine household activities: \textit{cleaning the window}, \textit{washing the dishes}, \textit{making a sandwich}, \textit{cooking a meal}, \textit{watching television}, \textit{reading a book}, \textit{doing laundry}, \textit{taking a shower}, and \textit{brushing teeth}.

\subsection{Demonstration: Real-World Robot}

While our evaluations use the VirtualHome simulator, we also deploy the system on a Stretch RE2 for real-world verification. The robot is capable of maintaining its own semantic map (albeit with substantial hardware limitations), identifying a person in the environment, and inferring their belief state. The core codebase and models were unchanged between the simulation and real-world deployments.

All code and data used in this work is available at: \url{https://github.com/gt-cec/tmm-hri}.

\begin{figure*}
    \centering
    \includegraphics[width=1.00\linewidth]{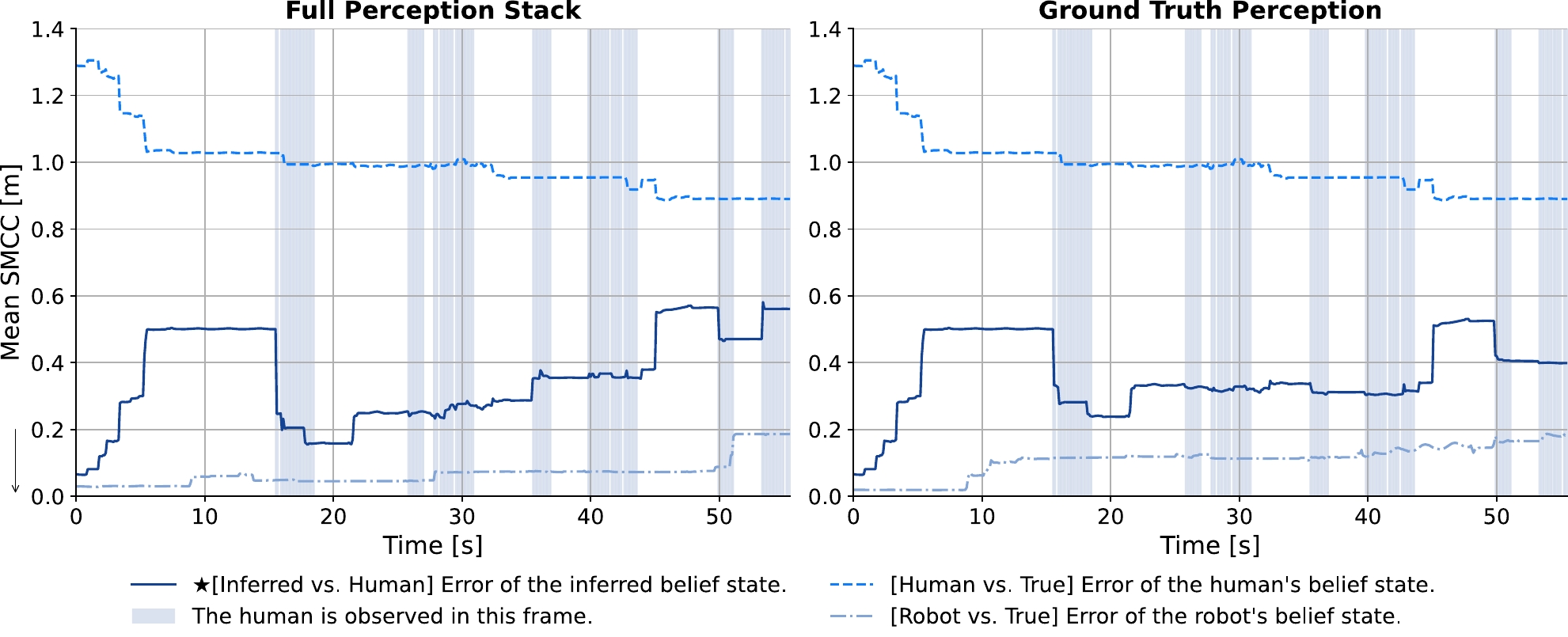}
    \caption{Performance of the belief state inference in an episode of the ``Parents are Out'' scenario. The scenario shuffles the environment, and then the human and robot walk through each room. The primary figure of merit is the error of the inferred belief state (indicated with $\star$), or the SMCC difference between the inferred belief state and the human agent's true belief state. Two trials are visualized -- \textbf{(left)} uses the robot's full perception stack (no privileged perception information) and \textbf{(right)} uses ground truth perception from the simulator. We find near-identical results, indicating an effective perception system.}
    \label{fig:parents_are_out}
\end{figure*}
\section{RESULTS}

We first evaluate simulator error using a static environment walkthrough. The simulator and belief states are initialized to the tidy household state, and the person and robot visit all rooms. Ideally, the belief states would negligibly change. However, VirtualHome does not provide a forward vector for agents nor the camera poses, causing a ``wavering'' effect as animations disconnect the estimated camera pose from the agent's forward vector, affecting object projection.

The static environment walkthrough resulted in a near-constant error of less than $0.25$m using the perception stack and $0.15$m using ground-truth perception across all three figures of merit. The value marks the portion of error that can be attributed to the simulator's limitations.

\subsection{Parents are Out Scenario Evaluation}

We then evaluate the \textit{``Parents are Out!''} scenario. Fig. \ref{fig:parents_are_out} shows the mean SMCC of each figure of merit for the scenario, comparing the perception stack and ground truth perception. Initially, the error of the person's belief state \(SMCC(\beta^{human}, \beta^{true})\) is high because they last recalled the household to be in the tidy state. As the person walks through the household, they observe objects and the error decreases. As the robot trails and observes the person and infers their belief state, its inference error \(SMCC(\widehat\beta^{human}, \beta^{human})\) fluctuates as it infers how the person would resolve their belief state given observations. 

\begin{table}[b!]
\caption{Ablation study on each perception module}
\centering
\begin{tabular}{ c c c|c }
    \shortstack{Object\\Detection} & \shortstack{Pose\\Detection} & \shortstack{Trajectory\\Inference} &
    \shortstack{Mean Inference\\Error [m]} \\
        \hline
        OWLv2 & RTM+MM & A* & 0.356 \\
        \textbf{GT} & RTM+MM & A* & 0.359 \\
        OWLv2 & \textbf{GT} & A* & 0.347 \\
        OWLv2 & RTM+MM & \textbf{GT} & 0.344 \\
        OWLv2 & RTM+MM & \textbf{None} & 0.345 \\
    \hline
        GT & GT & GT & 0.360 \\
        \textbf{OWLv2} & GT & GT & 0.378 \\
        GT & \textbf{RTM+MM} & GT & 0.376 \\
        GT & GT & \textbf{A*} & 0.378 \\
        GT & GT & \textbf{None} & 0.378 \\
    \hline
\end{tabular}
\label{table:ablation}
\end{table}

\subsection{Ablation Error Study}
Notably, Fig. \ref{fig:parents_are_out} shows near-identical plots for the perception stack and the ground truth perception. Ground truth perception used ground truth object detection \& segmentation, person poses, and person trajectories. The result indicates that our perception stack was highly accurate, with findings consistent across trials with other random seed initializations.

To identify sources of error we conduct an ablation study on each component of the perception stack. Table \ref{table:ablation} shows the mean inference error across all frames. We ablate each perception component relative to the perception stack and to ground truth perception from the simulator, and find only a marginal difference between the components. The remaining sources of error are the simulator itself, the belief state algorithm's resolver, and the consequences of operating in a partially-observable domain.

\subsection{Active Assistance Demonstration.}
The last evaluation demonstrates our methods in the active assistance task. Table \ref{table:active_assistance} details the performance of the demo at retrieving relevant objects for each activity across several points in the scenario. For consistency, the analysis used the same episode as in Fig. \ref{fig:parents_are_out}, resulting in a total of $207$ object relevance queries across all nine activities at six points in the episode. Each query was independently presented to each candidate semantic reasoning method without including the context from prior queries. The LLM prompts and encoder classification queries are available on the project webpage.



\begin{table}[b!]
\caption{Performance of semantic reasoning methods\\for the active assistance task}
\centering
\begin{tabular}{ c | c c c }
    Method & F1 & Precision & Recall \\
        \hline
        LLM (List CoT) & \textbf{0.68} & 0.56 & \textbf{0.86} \\
        LLM (Single CoT) & 0.53 & \textbf{0.77} & 0.41 \\
        LLM (Single Binary) & 0.46 & 0.76 & 0.33 \\
        deBERTa-v3 & 0.56 & 0.46 & 0.72 \\
        Random & 0.14 & 0.08 & 0.50 \\
    \hline
\end{tabular}
\label{table:active_assistance}
\end{table}

While all methods outperform random, there is no clear winner. In the LLM-based methods, the prompt variation resulted in dramatic effects on performance. From a user experience perspective, \textit{precision} is most important as it minimizes false positives and raises the value of the robot's recommendations. We anticipate that in real-world applications, people will prefer fewer recommendations than incorrect recommendations. Therefore, the single-object chain-of-thought prompt is most effective. Future work can explore a more rigorous agentic evaluation, potentially using an agent-as-a-judge verification, ensemble methods, tuning the chain-of-thought prompting, fine-tuning an encoder model for task-item classification, and using larger models.

\section{DISCUSSION}

Our methods succeeded at inferring a person's belief state by passively monitoring them in a household environment. Notably, in the simulation domain the perception stack performed nearly identical to the ground truth perception despite using zero-shot off-the-shelf models. Extending beyond perception, all modules made assumptions about human behavior and perception that can be relaxed in future work. 

The belief state algorithm performs remarkably well. As seen in Fig. \ref{fig:parents_are_out} with the \textit{Parents are Out!} scenario, simulator error \((\beta^{robot}, \beta^{true})\) accounted for more than a third of the inference error \((\widehat\beta^{human}, \beta^{human})\), and the inference error was substantially lower than the human's belief state error \((\beta^{human}, \beta^{true})\). However, the experiment assumed the human agent has perfect perception and resolves observed objects in the same manner as the robot. Future work can connect our methods with the human factors and human-robot interaction communities to improve the human model.

Modeling a person's perception and comprehension is an active research question that directly ties into the methods presented in this work. While this work models humans using ground truth perception, the binary ``object is seen'' approach can be replaced with a confidence value. This could be explored as a function of the object's angle relative to the person, or the duration for which the object is visible. Additionally, perception could be personalized using gaze data from an individual, such as by fine-tuning a saliency filter or training a more comprehensive model.

Fig. \ref{fig:parents_are_out} shows that the human's belief state error \((\beta^{human}, \beta^{true})\) does not converge to the simulator error \((\beta^{robot}, \beta^{true})\). This indicates the domain was challenging enough for the simulated person to not obtain a correct belief state after walking throughout the household. This characteristic enabled the \textit{active assistance} downstream task, and suggests that other household or floorplan domains would be sufficiently complex to cause a useful disparity between the robot's belief state and the person's belief state. 

The active assistance downstream task was successful at using semantic reasoning networks to identify objects to notify the person about. While all methods greatly outperformed random selection, the LLM-based methods had comparable performance to the encoder-only deBERTa model. In downstream applications, system designers should be mindful of the performance vs. computation tradeoffs of using general-purpose distilled LLMs for semantic reasoning. While an LLM approach was the highest performing in our demonstration (using \textit{precision} as our evaluation metric), the deBERTa approach had comparable if not superior performance in other metrics at a reduced computational cost.

\section{CONCLUSION \& FUTURE WORK}

In this work we formalize the problem of inferring a person's belief state (level one situation awareness) from third-party camera perspectives in a dynamic and partially-observable domain. We present a first approach to this problem, novel evaluation metrics, and demonstrate how an inferred belief state can be applied to downstream human-AI teaming tasks. We find that our approach can reason about human beliefs and we identify how future work can improve upon the presented architecture and implementation.

Team mental models have tremendous potential across a broad range of human-robot interaction and human-AI teaming domains. User intent recognition, proactive assistance, user behavior prediction, selective communication, and contextual communication can all leverage an inference of the user's belief state. However, advancements in this space are limited by the necessity of several key tools. 

Foremostly, future work would greatly benefit from a realistic simulator with an API for high-level point navigation, semantic object relocation, multiple agents, and access to instance-level and semantic-level RGB/D feeds. Existing simulators that advertise such capabilities (VirtualHome, Habitat 3.0) are challenged by bugs, low-to-no documentation, and extensive engineering requirements.

Future work can also explore new scenarios. One example is the sequel to ``Parents are Out!'' -- the ``Clean Up'' scenario where objects are continuously relocated by the agents to create a continuously dynamic domain. A second example is an industrial environment where the robot alerts a worker to hazardous objects that the worker is not aware of.

Lastly, no works have obtained a dataset of user situation awareness responses in a representative household or workplace environment. Such a dataset is needed to broadly align inferred belief states with actual user situation awareness. An ideal module could identify the probabilities of perceiving objects given the robot's understanding of the person's gaze history to relax the \textit{perfect perception} assumption herein.

Applying team mental models to human-robot teaming has incredible potential to augment current human-robot interaction capabilities and human-AI teaming. We are excited for the future of this field's in the robotics community.

\bibliographystyle{plain}
\bibliography{paper.bib}

\end{document}